\documentclass{ifacconf}

\usepackage{graphicx}      
\usepackage{graphics} 
\usepackage{natbib}        
\usepackage{amsmath,amssymb,amsfonts}
\usepackage{algorithmic,algorithm}
\usepackage{url}
\usepackage{siunitx}
\usepackage{xcolor}
\usepackage{multirow}
\usepackage{tabularray}

\usepackage{makecell} 
\usepackage{graphicx} 

\usepackage{svg}

\graphicspath{ {Figures/} }

\begin{document}
\begin{frontmatter}

\title{TRACE: \textit{T}raversability-aware \textit{R}eactive Navigation via VLM-driven \textit{A}daptive \textit{C}onstraints \textit{E}stimation} 

\thanks[footnoteinfo2]{This work has been partially funded by the European Union's Horizon Europe Research and Innovation Programme under the Grant Agreement No. 101119774 SPEAR, by the European Union through the Interreg Aurora programme (AUTONOMIC, NYPS-ID 20370377), and by the Bo Rydins Stiftelse för vetenskaplig forskning under the research project GAIA.}
\thanks[footnoteinfo4]{For a video of the article, showcasing the field experiments, please refer to the following link: \url{https://youtu.be/fG6C-aUE5I8}.}

\author[First]{Mario A.V. Saucedo}, 
\author[Second]{Midhun Edatharam Kunnath}, 
\author[First]{Elias Small}, 
\author[First]{Akash Patel}, 
\author[First]{Shruti Kotpalliwar}, 
\author[First]{Christoforos Kanellakis}, 
\author[First]{and George Nikolakopoulos}

\address[First]{Robotics and AI Team, Luleå University of Technology, Luleå, SE-97187, Sweden. Corresponding author`s email: marval@ltu.se.}
\address[Second]{Korea Advanced Institute of Science \& Technology (KAIST), Daejeon, 34141, South Korea.}

\begin{abstract}                
This paper presents TRACE, a traversability-aware reactive navigation framework designed for the autonomous operation of ground robots in unstructured forest environments. 
Unlike traditional approaches, TRACE leverages VLM-based confidence maps from positive and negative natural-language prompt sets and combines them through evidential fusion to produce a dense traversability map that prioritizes the terrain based on signs of previously traversed paths. 
Traversability-annotated points are then extracted and used to derive the left and right boundaries of the traversable path via percentile binning followed by spline or regularized polynomial fitting. These boundaries are then incorporated as soft barrier constraints within an NMPC formulation, enabling safe and adaptive path following in complex natural environments. 
Rigorous experimental trials carried out in real-life forest environments demonstrate TRACE ability for smooth navigation in cluttered, deformable, and visually ambiguous terrains. 

\end{abstract}


\end{frontmatter}

    \begin{figure*}[!htbp]
        \centering
        \includegraphics[width=\textwidth]{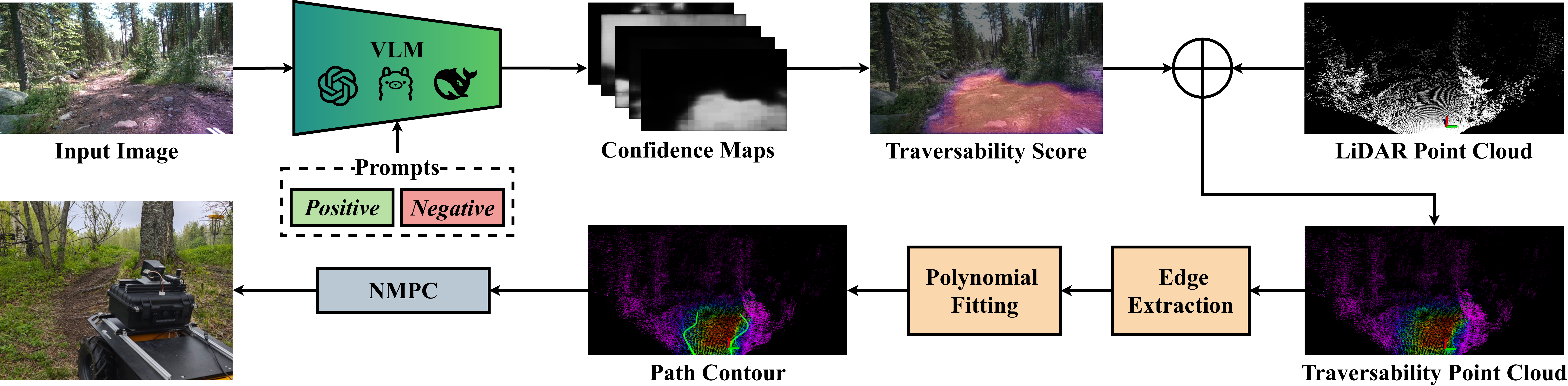}
        \caption{\textbf{Framework Overview.} Block diagram of the proposed framework, where VLM-based confidence maps from positive and negative prompt sets are combined through evidential fusion to produce a dense traversability map. The map is then projected onto 3D to obtain traversability-annotated points, which are later used to extract the left and right traversable boundaries based on percentile binning and spline/regularized polynomial fitting. Finally, the resulting boundaries are incorporated as soft barrier constraints within the NMPC formulation. 
        }
        \label{fig:pro}
    \end{figure*}

\section{Introduction}


Autonomous robotic navigation in forest environments plays a crucial role in numerous domains, including precision forestry management, ecological monitoring, biodiversity assessment, and disaster response operations such as search and rescue or wildfire inspection. Despite its importance, reliable ground robot navigation in such environments remains a major challenge. Forest terrains are inherently unstructured and highly variable, characterized by uneven ground surfaces, loose soil, roots, dense undergrowth, and natural obstacles like fallen logs, rocks, and thick vegetation. 
A central difficulty lies in accurately identifying traversable regions within this complex environment. Traditional geometry-based methods, such as those relying solely on point cloud elevation, slope, or roughness, often struggle to interpret the rich visual and semantic complexity of natural scenes. 
While semantic segmentation methods 
often fail to capture the functional traversability of the environment, that is, whether a region is truly traversable, since appearance or semantics alone do not always correspond to physical feasibility (\cite{2022survey}). For instance, both moss-covered ground and dense grass may share similar visual features yet differ drastically in their ability to support the robot’s movement. Moreover, forest environments exhibit dynamic lighting, seasonal variations, and occlusions from vegetation, all of which can degrade segmentation accuracy and lead to inconsistent terrain interpretation.

To overcome these limitations, a more human-inspired approach to navigation is required. 
Humans navigating through forests rarely rely solely on recognizing semantic categories like “dirt” or “vegetation.” Instead, we intuitively follow loosely structured paths, such as worn trails, clearings between trees, or subtle traces of previous traversal visible on the ground. These cues encode valuable information about accessibility and safety, formed by the natural flow of movement through the environment.
Vision–Language Models (VLMs) provide a powerful foundation for this human-like perception by linking visual understanding with high-level semantic and contextual reasoning (\cite{VLMs}). Unlike conventional segmentation or geometric-based methods, VLMs are trained on vast amounts of image–text pairs, enabling them to associate visual patterns, such as textures, shapes, and spatial layouts, with abstract linguistic concepts like “trail,” “open space,” or “obstacle.” 
For instance, a VLM can recognize subtle cues such as faint ground traces, vegetation bending, or lighting gradients as indicators of a traversable path, much like how humans intuitively interpret these cues when moving through forests.
By emulating this behavior, autonomous systems can perceive and reason about navigational affordances, that is, the implicit pathways, patterns, and textures that suggest traversability.


In this work, we propose a navigation framework that leverages VLMs to enable human-like reasoning for traver-sability assessment in forest environments. Rather than relying solely on predefined geometric thresholds or rigid semantic segmentation, our system interprets the scene through contextual, language-guided perception, allowing it to identify implicit trails, clearings, and accessible regions in a manner similar to human navigation. By combining this high-level semantic reasoning with a nonlinear model predictive control (NMPC), 
the approach enables adaptive path discovery and robust motion planning in highly unstructured and cluttered terrains. An overview of our approach is presented in Fig.~\ref{fig:pro}.



\subsection{Contributions}
The main contributions of this work are threefold. First, we introduce TRACE, a traversability-aware reactive navigation framework that unifies vision–language perception, constraint estimation, and nonlinear model predictive control (NMPC) for autonomous navigation in unstructured forest environments. The proposed approach leverages Vision–Language Models (VLMs) with evidential fusion on positive and negative natural-language prompt sets to infer dense traversability maps, enabling a human-like understanding of terrain affordances that transcends explicit semantic or geometric cues. Second, we develop an adaptive contour extraction pipeline based on percentile-driven boundary detection and regularized polynomial fitting, which converts traversability-annotated point clouds into smooth, continuous representations of the navigable corridor. These contours are integrated as soft barrier constraints within the NMPC formulation to ensure safe, dynamically feasible motion through complex natural environments. Third, we present an extensive experimental validation of the framework in real-world forest environments in northern Sweden using a Husky mobile robot. The results of the field test demonstrate TRACE's robust capabilities to generalize across varying environments while maintaining safe navigation in cluttered, deformable, and visually ambiguous terrains.




\section{Related Work}
\textbf{Classical Traversability Analysis and Learning-Based Approaches.} Terrain traversability analysis has been extensively addressed in the context of autonomous ground vehicles operating in unstructured outdoor environments (\cite{2022survey}). 
Classical geometric-based approaches primarily estimate traversability from terrain morphology by evaluating slope, roughness, or discontinuities derived from LiDAR or stereo data. Although computationally efficient, these methods often underperform in vegetated or deformable terrains, where visual appearance is not a reliable indicator of physical traversability. To enhance robustness, multimodal perception systems have been proposed, such as the fusion of LiDAR and camera data for terrain mapping and the combination of geometric and visual cues for terrain classification. The introduction of semantic segmentation further improved environmental understanding by providing categorical terrain labels. However, such models remain sensitive to illumination changes, occlusions, and domain shifts between training and deployment environments.

To address these generalization issues, learning-based methods have been introduced. 
\cite{rev1} introduced a domain adaptation framework with a safety-preserving loss to maintain reliable traversability estimation across unseen domains. \cite{wayfast} proposed a predictive learning model for dynamic traversability estimation, while \cite{ganav} achieved efficient terrain segmentation in outdoor environments using lightweight neural architectures. \cite{excavator} presented the Terrain Navigability System (TNS) for autonomous excavators, integrating mapping and control for operation on deformable terrain. \cite{saucedo2023eat} later introduced Environment-Agnostic Traversability\\ (EAT), which emphasizes reactive navigation through models capable of generalizing across diverse terrains. Despite these advances, most classical and learning-based approaches rely on explicit geometric or semantic labels, limiting their ability to infer contextual or affordance-level cues that are essential for reasoning about traversability.

\textbf{VLM-based and Context-Aware Traversability.} Recent works have demonstrated the potential of VLMs for context-aware navigation and affordance reasoning. 
\cite{adarsh2024} employed VLMs for context-aware navigation in both indoor and outdoor environments. \cite{mohamed2025} integrated physically grounded vision–language representations for outdoor traversal reasoning. Similarly, \cite{zhen2024} demonstrated the use of large language and vision–language models for interactive navigation among traversable obstacles. \cite{daeun2025} developed a trajectory generation and selection system that leverages VLMs for semantic reasoning in mapless outdoor navigation. More recently, \cite{gummadi2025} proposed a zero-shot traversability framework for unknown environments, while \cite{sahu2025} combined VLM-based perception with human-in-the-loop supervision for enhanced adaptability in off-road conditions. 

In contrast to prior approaches, this work explores a human-inspired navigation framework that leverages VLMs to infer traversability through implicit environmental cues, such as loosely structured trails and surface traces, enabling robust, context-aware navigation in forest environments beyond the limits of purely geometric or semantic methods.

\section{Problem Formulation}

In this article, we seek an optimal navigation strategy for an autonomous robotic agent operating in a forest environment by dynamically inferring a traversable corridor via VLMs.
These boundaries are then represented as two bounding polynomials and imposed as barrier constraints within a nonlinear model predictive control to ensure safe, smooth, and feasible motion.

\textbf{Robot Kinematic Model.} 
The robot is represented using the kinematic model proposed by \cite{model}, 
%
where the control inputs $\mathbf{u}$ are the commanded linear $v$ and angular $\omega$ velocities, while the states $\mathbf{x}$ are the $(x,y) \in \mathbb{R}^2$ position of the geometric center and, $\theta$ the heading angle.



\textbf{Traversability Constraints.} 
The traversable region in a local ground-aligned frame 
is approximated by two $d$-degree polynomials:
\begin{equation}
p_L(x) = \sum_{i=0}^{d} c_{L,i}\, x^{i},\,\,
p_R(x) = \sum_{i=0}^{d} c_{R,i}\, x^{i},
\end{equation}
representing the left and right boundaries of the corridor, where $C_L$ and $C_R$ are the polynomial coefficients.
The admissible lateral region is defined as
\begin{equation}
y \in [\,p_L(x)+m,\; p_R(x)-m\,],
\end{equation}
where $m>0$ is a safety margin.

\section{Methodology}


\textbf{VLM-based Traversability Scoring. }
Given an input image \( I \in \mathbb{R}^{H \times W \times 3} \), a VLM model produces dense prompt-conditioned confidence maps that quantify the correspondence between visual regions and textual descriptions. In this work, two sets of prompts are defined: a \emph{positive set} $\mathcal{P}=\{p_1,\ldots,p_{i}\}$ representing traversable terrain concepts (e.g., \textit{``trail''}, \textit{``open ground''}, \textit{``clear path''}), and a \emph{negative set} $\mathcal{N}=\{n_1,\ldots,n_{j}\}$ representing obstacles or non-traversable terrain (e.g., \textit{``fallen log''}, \textit{``dense vegetation''}, \textit{``rock''}). For each prompt, the model outputs a confidence map $R\in[0,1]^{H \times W}$, 
interpreted as \emph{evidence} \(e_p = R_p\) supporting its corresponding class.
%
%
%
The total positive and negative evidence at each pixel \((h,w)\) are obtained by: 
\begin{equation}
\begin{aligned}
\alpha^{+}(h,w) &= 1 + \sum_{p\in\mathcal{P}} e_p(h,w), \\
\alpha^{-}(h,w) &= 1 + \sum_{n\in\mathcal{N}} e_n(h,w).
\end{aligned}
\end{equation}
Assuming a Beta-distributed belief over traversability, we apply evidential fusion (\cite{thrun2005probabilistic}) and obtain the expected probability for a pixel being traversable as
\begin{equation}
E(h,w) = \frac{\alpha^{+}(h,w)}{\alpha^{+}(h,w) + \alpha^{-}(h,w)},
\end{equation}
with uncertainty
\begin{equation}
U(h,w) = 2/(\alpha^{+}(h,w)+\alpha^{-}(h,w)).
\end{equation}
The resulting traversability map is defined as:
\begin{equation}
C(h,w) = E(h,w)\,\bigl(1 - U(h,w)\bigr), \qquad C \in [0,1],
\end{equation}
where higher values indicate higher traversability likelihood and lower values correspond to obstacle-like regions.  




Finally, the traversability map $C{(h,w)}$ is then integrated with the point cloud information of a depth sensor, expressed in a local coordinate frame corresponding to the $X$-axis forward, $Y$-axis towards the left, and $Z$-axis upwards of the robot. By reprojecting each 3D point $\mathbf{p} = (x, y, z)$ into the image plane using the camera intrinsics and extrinsics parameters, yielding pixel coordinates $(i, j)$. The corresponding traversability score $c = C{(h,w)}$ is then sampled from the confidence map and assigned to the point. This produces an augmented point cloud $\mathcal{D} = \{(\mathbf{p},\, c)\}$, where each point encodes both position and traversability score. 


\textbf{Percentile-based Edge Extraction. }
To isolate traver- sable ground regions, points are first filtered according to their traversability score. Let $\theta_c$ denote a confidence threshold. The retained subset is defined as
\begin{equation}
\hat{\mathcal{D}} = \{(\mathbf{p},\,\,c)\in\mathcal{D}\mid \, c \ge \theta_c\}.
\end{equation}
Where the $X$-axis is constrained to $R_{\max}$ meters to prevent extrapolation from sparse measurements.

Next, to identify the lateral limits of the traversable region, the filtered points are divided into $B$ bins along the $X$-axis. For each bin $b \in B$, the lateral coordinates $y$ of all points within that bin are collected, and the left and right boundaries are estimated using percentile statistics:
\begin{equation}
y^{\mathrm{L}}_b = \operatorname{perc}_{p_L}(y_b), \qquad
y^{\mathrm{R}}_b = \operatorname{perc}_{p_R}(y_b),
\end{equation}
where $p_L$ and $p_R$ denote the lower and upper percentiles (i.e., $5\%$ and $95\%$). Each bin is assigned a representative $\bar{x}_b$ coordinate and a $w_b$ confidence weight proportional to the bin’s population and median confidence. These discrete boundary samples are smoothed with an edge-preserving moving average and tapered at the endpoints to mitigate quantile jitter and boundary oscillations.

\textbf{Polynomial Edge Fitting. }
Finally, the smoothed boundary samples $\{(\bar{x}_b, y^{\mathrm{L}}_b, w_b)\}$ and $\{(\bar{x}_b, y^{\mathrm{R}}_b, w_b)\}$ are fitted using a regularized polynomial or spline regression to obtain continuous edge functions $p_{L}(x)$ and $p_{R}(x)$. A smoothing factor $\lambda_s > 0$ controls the trade-off between data fidelity and smoothness, where larger values enforce smoother edges at the cost of precision. The fitting procedure follows a hierarchical strategy:
\begin{enumerate}
    \item Cubic smoothing spline with smoothing parameter $s=\lambda_s B\,\mathrm{Var}(y)$.
    \item Weighted Chebyshev ridge regression if spline fitting is unstable.
    \item Weighted monomial ridge regression as a fallback.
\end{enumerate}
The resulting functions are densely sampled within the supported forward range to generate smooth boundary curves representing the left and right edges of the traversable corridor. 

\section{Experimental Evaluation}


In this section, we first compare the performance of our method with state-of-the-art approaches and then present rigorous experimental validations conducted in real forest environments. We use CLIPSeg (\cite{clip}) as the visual–language model, selected for its ability to provide zero-shot segmentation from text prompts while maintaining real-time performance.

\subsection{Performance Comparison}

\begin{figure}[!htbp]
        \centering
        \includegraphics[width=\linewidth]{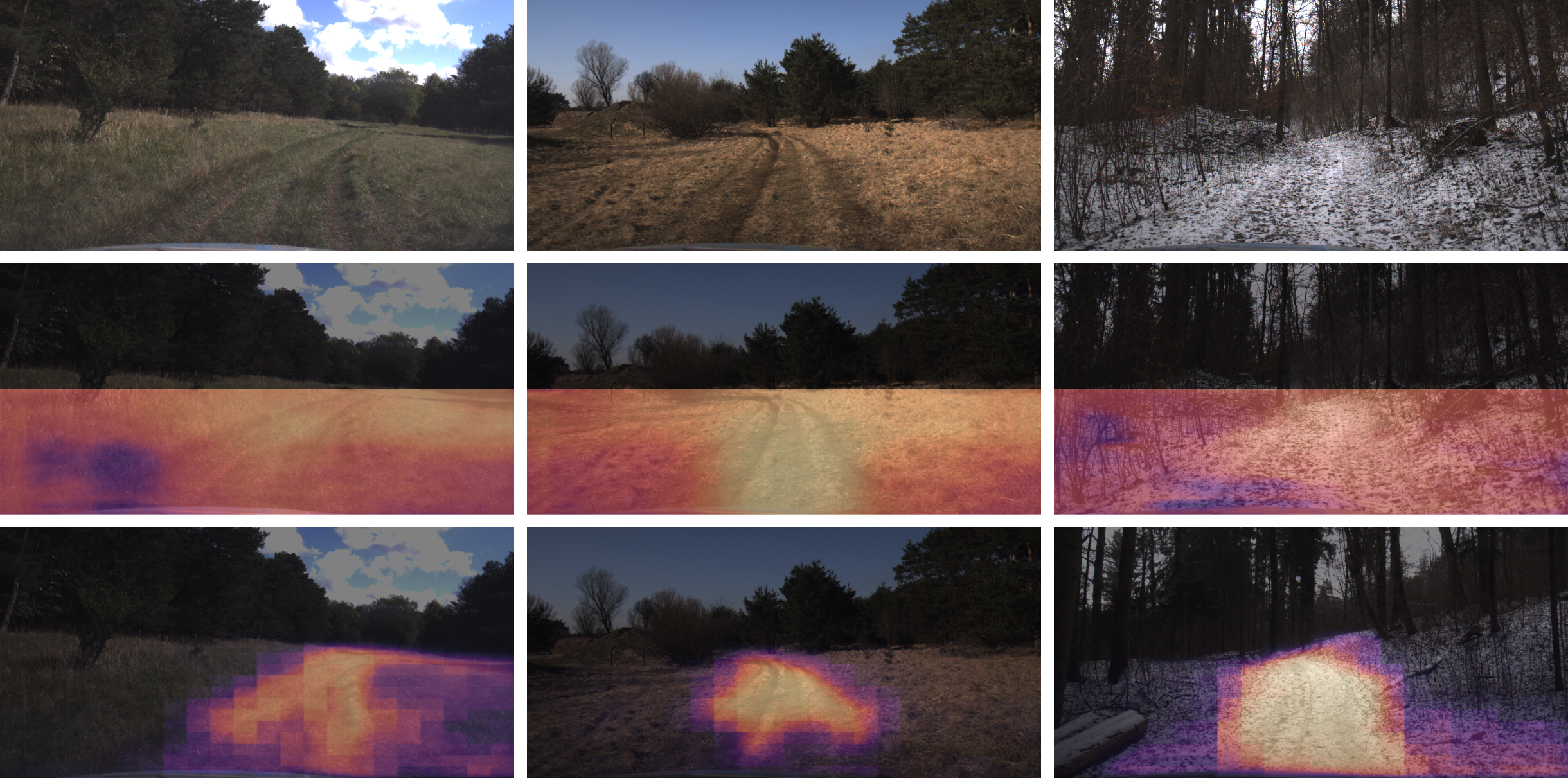}
        \caption{\textbf{Traversability Maps.} Visualizations of the estimated traversability maps on different challenging samples from the GOOSE dataset (\cite{goose}), illustrating the comparison between the baseline's (i.e., middle row) (\cite{wayfast}) and the proposed method’s (i.e., bottom row) ability to adapt to different environmental conditions.}
        \label{fig:tra}
    \end{figure}

\textbf{Traversability Estimation.} 
%
The performance of the proposed method for traversability scoring is first compared against the baseline for self-supervised traversable path estimation presented in \cite{wayfast}.
To ensure consistency and fairness with the data use during the training process of the baseline, only the bottom half of the images was considered during testing, corresponding to the baseline’s lower field of view (FOV).
The comparison was done using the GOOSE dataset (\cite{goose}), consisting of multiple sequences of unstructured off-road environments under varying conditions, including different seasons presenting dry vegetation and snow. 
%
%
Figure \ref{fig:tra} demonstrates the proposed framework's advantages over the baseline method (\cite{wayfast}), on different challenging samples from the dataset. 
Specifically, the proposed method yields sharper and more localized traversability activations that accurately align with the drivable regions, rather than exhibiting the diffuse responses observed in the baseline, resulting in improved spatial precision and boundary awareness. 
Moreover, the proposed approach exhibits enhanced robustness and interpretability, providing more reliable and semantically consistent confidence maps under varying environmental conditions.

The advantages of the proposed method are further illustrated when comparing the estimated traversability scores against human-defined ground-truth labels.
Since the current state of the art does not contemplate human-annotated datasets for path estimation in unstructured environments, we have opted to instead use the ground-truth labels for terrain classification based on navigability levels provided in \cite{ganav}. 
Table~\ref{table:miou} presents the results for intersection-over-union (IoU) between the traversability estimated with the proposed method and the ground-truth labels provided in \cite{ganav} for the GOOSE dataset (\cite{goose}). 
Where the \textit{Stable} class (i.e. the most optimal terrain) presents the highest intersection-over-union, followed by the \textit{Granular} class (i.e. the second most optimal terrain), demonstrating a high degree of consistency between the proposed method and the ground-truth annotations.



\begin{table}[h!]
\centering
\caption{IoU for the estimated terrain traversability on the \cite{ganav} labels.}
\label{table:miou}
\resizebox{\linewidth}{!}{
\begin{tabular}{c|ccccc}
\hline
\textbf{Label} & 
\makecell{\textbf{Stable}} & 
\makecell{\textbf{Granular}} & 
\makecell{\textbf{Poor}\\\textbf{foothold}} & 
\makecell{\textbf{High}\\\textbf{resistance}} & 
\makecell{\textbf{Obstacle}} \\ \hline \hline
IoU & 0.672421 & 0.334350 & 0.070665 & 0.007723 & 0.001812 \\ \hline
\end{tabular}
}
\end{table}

    \begin{figure}[!b]
        \centering
        \includegraphics[width=\linewidth]{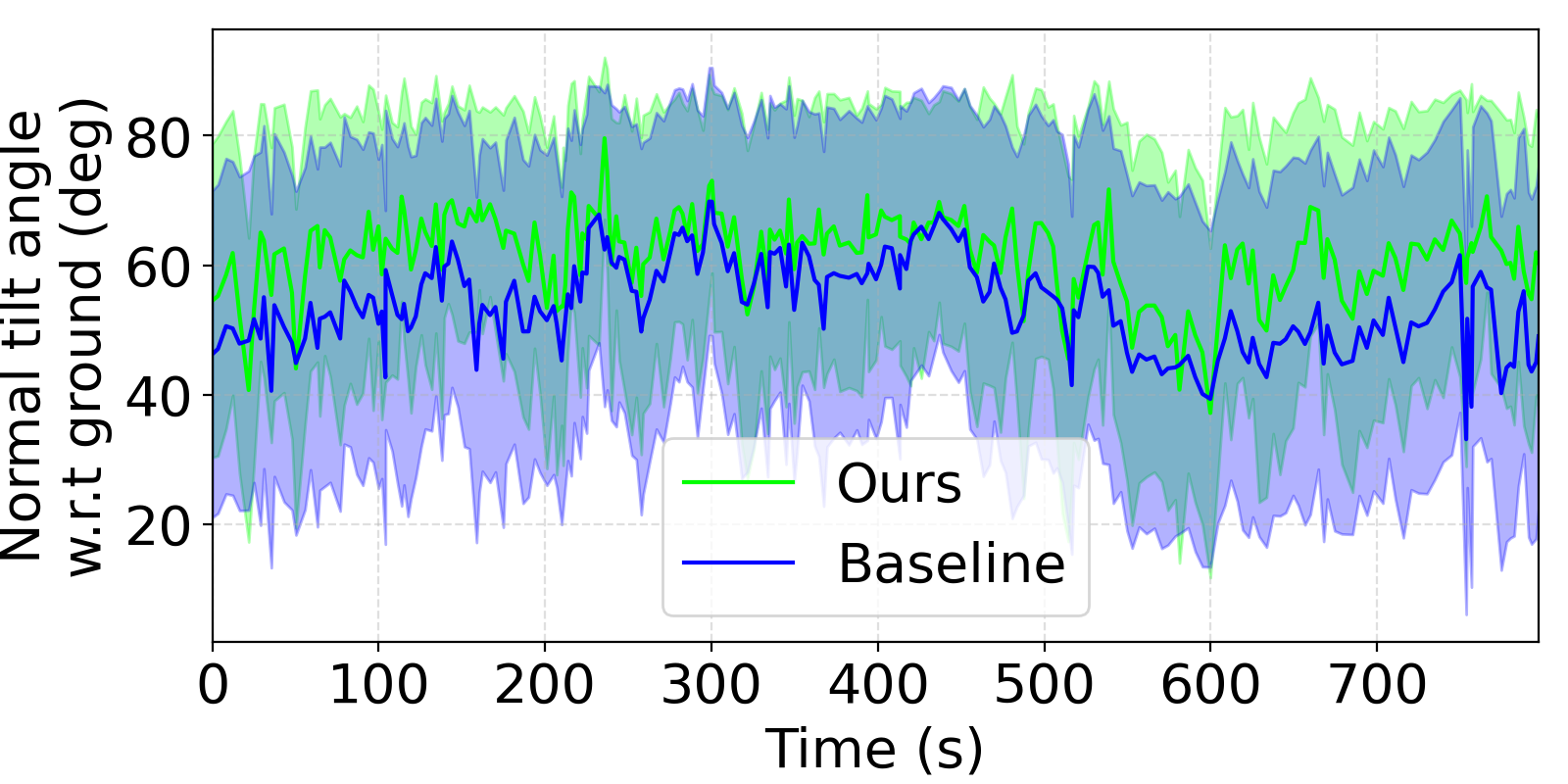}
        \caption{\textbf{Normal Tilt Angle.} Average surface-normal tilt angle for the traversable area estimated by the proposed semantic-based method compared to the LiDAR-only baseline. 
        }
        \label{fig:plot}
    \end{figure}

    \begin{figure*}[!htbp]
        \centering
        \includegraphics[width=\textwidth]{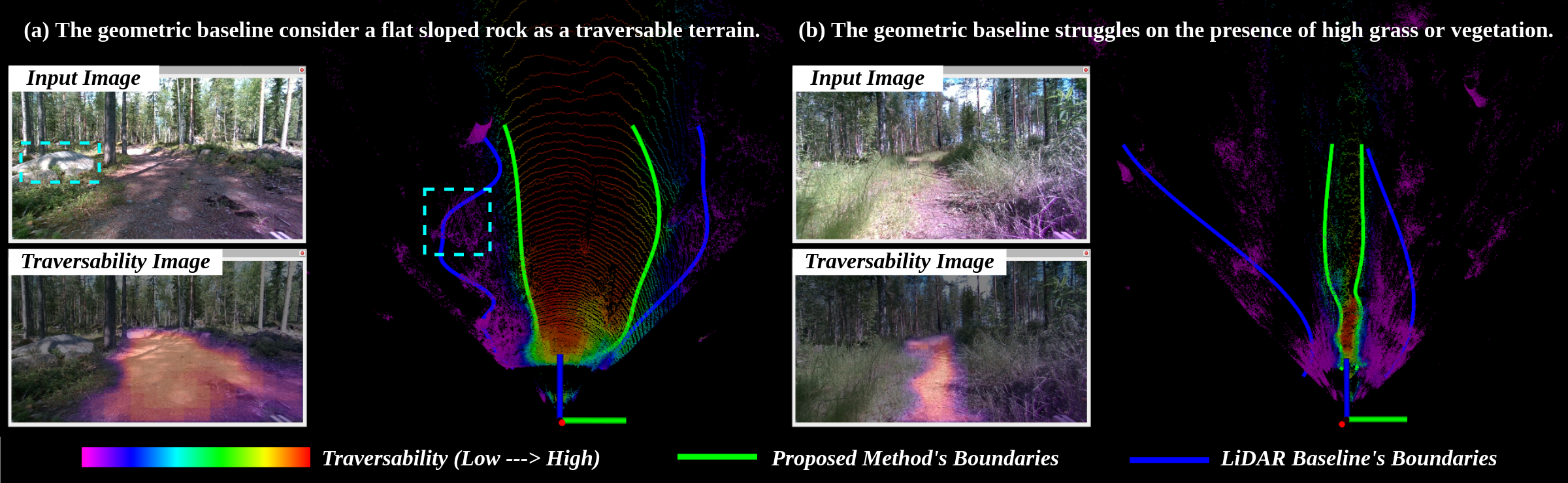}
        \caption{\textbf{Path Boundaries.} Comparison of path boundary estimation between LiDAR-only baseline and the proposed approach, where the baseline misclassifies certain sloped or vegetated areas as traversable, while TRACE correctly identifies them as unsafe and extracts smooth, coherent boundaries around the truly navigable regions.}
        \label{fig:comp}
    \end{figure*}

\textbf{Contour Estimation.} 
%
The contour estimation performance is assessed through a comparison between the proposed framework and a baseline that relies solely on geometric information. 
Specifically, the baseline method was developed following a traditional LiDAR-based approach, where boundary points are extracted by identifying discontinuities in the range profile, and then fitting a polynomial curve to these boundary regions to represent the left and right limits of the traversable area.
Figure \ref{fig:plot} compares the average surface-normal tilt angle with respect to the ground plane $(XY)$, for the area contained between the traversable boundaries estimated by the proposed VLM-based method against the geometric LiDAR-only baseline. The results show that the LiDAR baseline exhibits higher standard deviation in the tilt angle, reflecting its sensitivity to range discontinuities, occlusions, and sparse returns, causing unstable or jagged boundary estimates. In contrast, the proposed framework, despite incorporating only visual sensor information for the computation of traversability, produces a smoother and more consistent tilt profile, especially in cluttered or visually ambiguous environments where geometric cues alone become noisy.

Moreover, Fig.~\ref{fig:comp} illustrates two representative failure cases of the LiDAR baseline versus the proposed vision–language approach. In (a), the geometric baseline incorrectly interprets a sloped rock surface as traversable, whereas the proposed method accurately recognizes it as unsafe. In (b), the baseline fails to delineate boundaries in the presence of dense grass and vegetation, while the proposed model successfully infers a clear, continuous traversable path.
These examples further highlight the advantages of the proposed framework for contextual understanding and environmental robustness, and showcase its ability for reliable navigation in complex natural terrains.





\subsection{Field Test}

\begin{figure}[!h]
        \centering
        \includegraphics[width=\linewidth]{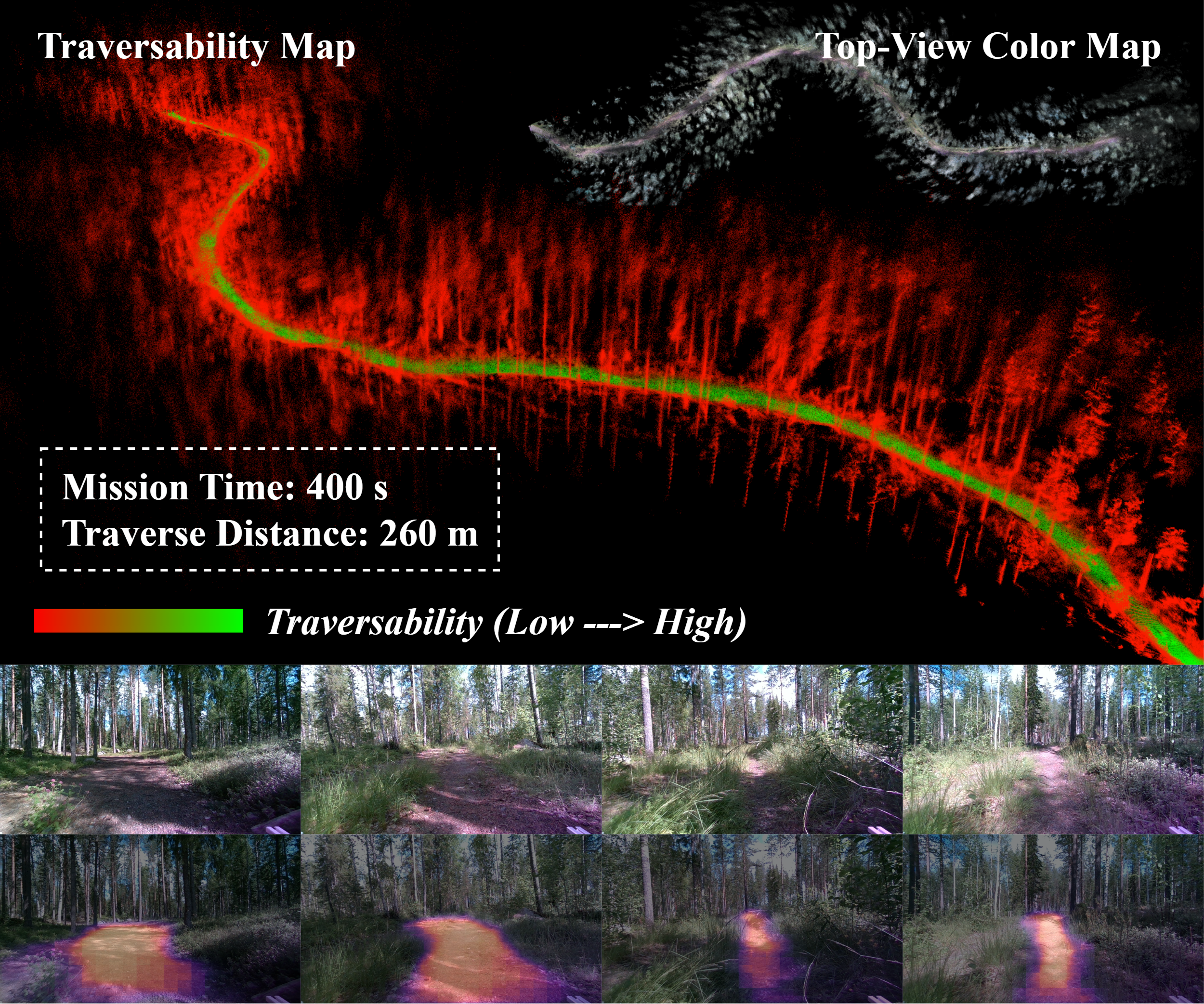}
        \vspace{-15pt}
        \caption{\textbf{Exp. 1.} Forest road obstructed by overgrow vegetation.
        }
        \label{fig:res1}
        \vspace{-1.4em}
    \end{figure}

\begin{figure}[!b]
        \centering
        \includegraphics[width=\linewidth]{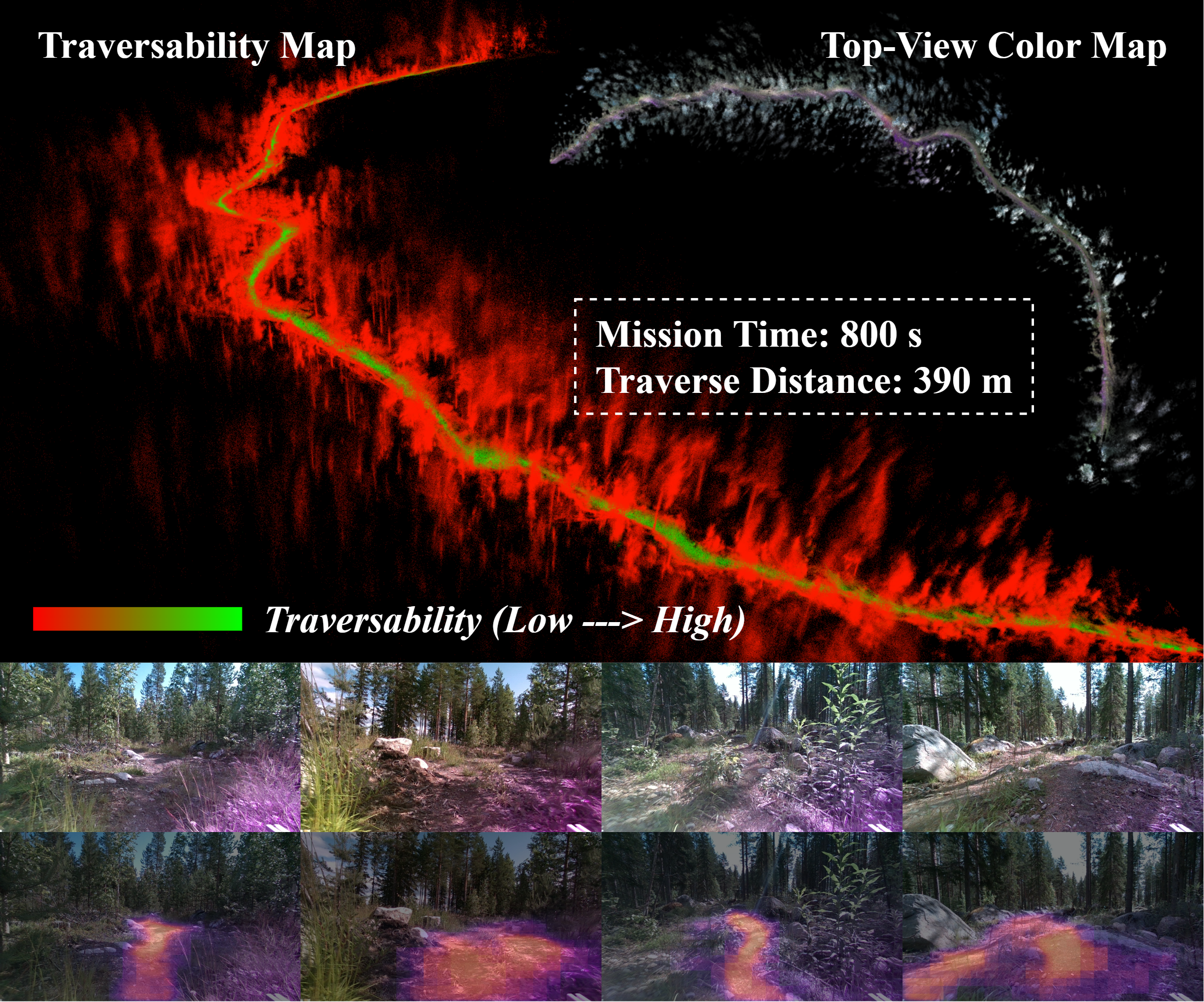}
        \vspace{-15pt}
        \caption{\textbf{Exp. 2.} Narrow hiking trail surrounded by rocks and vegetation.
        }
        \label{fig:res2}
    \end{figure}

The experiments took place in a real-life forest environment in northern Sweden with a Husky mobile robot. The objective of the experiments was for the robot to follow the loosely defined forest roads and hiking trails present in the environment. 
In the first experiment, showcased in Fig.~\ref{fig:res1}, the robot was made to navigate a forest road with multiple sections of tall grass and overgrown vegetation, which limited the visibility of the tire tracks left by previously transiting vehicles. On the other hand, the second experiment depicted in Fig.~\ref{fig:res2} took place in a hiking trail consisting of narrow passages across rocks and vegetation, presenting multiple slopes and meandering across the way. 
Both figures present at the top a visualization of the estimated traversability map, alongside a top-view of the corresponding color map, while at the bottom, images of the estimated traversability scores at different challenging scenes can be found. 
Overall, these experiments demonstrate the ability of the proposed framework to imitate human-like behaviors while traversing unstructured forest environments by tracing the loosely defined signs of previously traversed paths.










\section{Conclusion}

This work introduced TRACE, a traversability-aware navigation framework that integrates Vision–Language perception with adaptive contour estimation for reactive motion planning in forest environments. By exploiting the contextual understanding of VLMs, the framework bridges the gap between semantic perception and functional affordance reasoning, enabling robots to interpret implicit environmental cues, such as worn trails and subtle surface continuity, beyond explicit geometric or categorical labels. The proposed VLM-driven traversability maps in conjunction with percentile-based boundary extraction and polynomial contour fitting allow for smooth and stable NMPC integration, ensuring safe navigation in cluttered, deformable, and visually ambiguous terrains. Real-world experiments confirmed the system’s capability to adapt to different environmental conditions, achieving high spatial consistency and stable motion in unstructured trails. Future work will focus on extending the experimental evaluation to a wider range of environments, and deploying TRACE across a wider range of robotic platforms, such as legged robots and humanoid robots.




\section*{DECLARATION OF GENERATIVE AI AND AI-ASSISTED TECHNOLOGIES IN THE WRITING PROCESS}\vspace{-0.5em}
During the preparation of this work, the author(s) used the ChatGPT model developed by OpenAI, in order to correct grammar and refine the wording of the manuscript. After using this tool/service, the author(s) reviewed and edited the content as needed and take(s) full responsibility for the content of the publication.


\bibliography{ifacconf}             
                                                   







\end{document}